\documentclass[conference]{IEEEtran}
\IEEEoverridecommandlockouts
\usepackage{cite}
\usepackage{amsmath,amssymb,amsfonts}
\usepackage{algorithmic}
\usepackage{graphicx}
\usepackage{textcomp}
\usepackage{xcolor}
\usepackage{pifont}
\usepackage{amssymb}
\usepackage{enumitem}
\usepackage{multirow}
\usepackage{array}
\usepackage{subcaption}
\usepackage{times}
\usepackage{soul}
\usepackage{url}
\usepackage{amsmath}
\usepackage{amsthm}
\usepackage{booktabs}
\usepackage{algorithm}
\usepackage{algorithmic}
\usepackage[switch]{lineno}
\usepackage{colortbl}

\usepackage[most]{tcolorbox}
\newtcolorbox{promptbox}{
colback=gray!5,  % Background color
colframe=black!75, % Frame color
left=1em, % Padding left
right=1em, % Padding right
top=1em, % Padding top
bottom=1em, % Padding bottom
sharp corners, % Square corners
boxrule=1pt % Border thickness
}

\usepackage{newfloat}
\usepackage{listings}

\def\BibTeX{{\rm B\kern-.05em{\sc i\kern-.025em b}\kern-.08em
    T\kern-.1667em\lower.7ex\hbox{E}\kern-.125emX}}
\begin{document}

\title{Catching Chameleons: \\Detecting Evolving Disinformation\\ Generated using Large Language Models}
% \title{Exploring the Evolving Nature of \\LLM-Generated Disinformation}
% \thanks{*These authors contributed equally.}

% \author{\IEEEauthorblockN{Anonymous Authors}}
\author{\IEEEauthorblockN{Bohan Jiang*, Chengshuai Zhao*, Zhen Tan, Huan Liu}
\IEEEauthorblockA{\textit{School of Computing and Augmented Intelligence} \\
\textit{Arizona State University, USA}\\
\{bjiang14, czhao93, ztan36, huanliu\}@asu.edu}
% \and
% \IEEEauthorblockN{2\textsuperscript{nd} Chengshuai Zhao*}
% \IEEEauthorblockA{\textit{School of Computing and AI} \\
% \textit{Arizona State University}\\
% % City, Country \\
% czhao93@asu.edu}
% \and 
% \IEEEauthorblockN{3\textsuperscript{rd} Zhen Tan}
% \IEEEauthorblockA{\textit{School of Computing and AI} \\
% \textit{Arizona State University}\\
% % City, Country \\
% ztan36@asu.edu}
% \and 
% \IEEEauthorblockN{4\textsuperscript{th} Huan Liu}
% \IEEEauthorblockA{\textit{School of Computing and AI} \\
% \textit{Arizona State University}\\
% % City, Country \\
% huanliu@asu.edu}
}

\maketitle

\begingroup\renewcommand\thefootnote{\fnsymbol{footnote}}
\footnotetext[1]{These authors contributed equally.}
\endgroup

\begin{abstract}
Despite recent advancements in detecting disinformation generated by large language models (LLMs), current efforts overlook the ever-evolving nature of this disinformation. In this work, we investigate a challenging yet practical research problem of \textit{detecting evolving LLM-generated disinformation.} Disinformation evolves constantly through the rapid development of LLMs and their variants. As a consequence, the detection model faces significant challenges. First, it is inefficient to train separate models for each disinformation generator. Second, the performance decreases in scenarios when evolving LLM-generated disinformation is encountered in sequential order. To address this problem, we propose \textit{DELD} (\underline{D}etecting \underline{E}volving \underline{L}LM-generated \underline{D}isinformation), a parameter-efficient approach that jointly leverages the general fact-checking capabilities of pre-trained language models (PLM) and the independent disinformation generation characteristics of various LLMs. In particular, the learned characteristics are concatenated sequentially to facilitate knowledge accumulation and transformation. \textit{DELD} addresses the issue of label scarcity by integrating the semantic embeddings of disinformation with trainable soft prompts to elicit model-specific knowledge. Our experiments show that \textit{DELD} significantly outperforms state-of-the-art methods. Moreover, our method provides critical insights into the unique patterns of disinformation generation across different LLMs, offering valuable perspectives in this line of research.
\end{abstract}

\begin{IEEEkeywords}
disinformation, large language model, prompt learning
\end{IEEEkeywords}

\section{Introduction}
The emergence of large language models (LLMs) has fundamentally transformed the paradigm of online content generation~\cite{tan2024large}. Models like ChatGPT~\cite{OpenAI2023GPT4TR} and LLaMA~\cite{touvron2023llama} are now capable of composing human-level social media posts, news stories, and fiction~\cite{zhao2023survey,jiang2024media}. These advancements have democratized content creation, enabling unprecedented levels of creativity and efficiency. However, they have also supercharged the threat of disinformation worldwide~\cite{chen2023can}. Disinformation, defined as intentionally false or misleading information, poses severe risks to public trust and the integrity of information ecosystems~\cite{shu2020disinformation}. The ease with which anyone can now generate fake but convincing content exacerbates these risks. For example, a malicious actor could use an LLM to generate a false news article about a health crisis, spreading panic and disinformation~\cite{zhou2023synthetic}. The rapid dissemination of such false information on digital platforms can outpace fact-checking efforts, leading to widespread disinformation before accurate information can correct the narrative~\cite{spitale2023ai}. This rapid spread is particularly problematic in times of crisis, where timely and accurate information is critical.

\begin{figure}[!t]
\centering
\includegraphics[width=1\columnwidth]{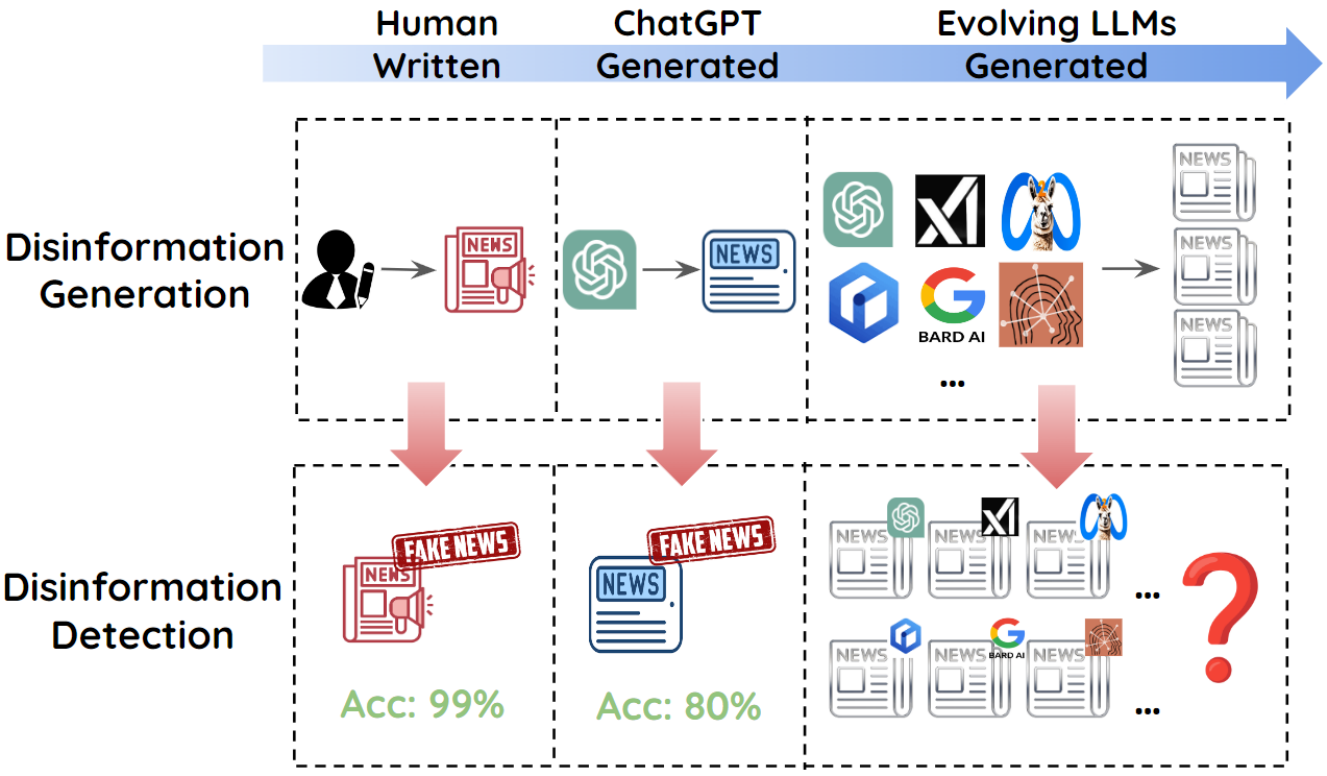} 
% \vspace{-.2cm}
\caption{An overview of online disinformation generation and detection pathway over time. During the post-LLM era, we mainly focused on detecting human-written disinformation. After the launch of ChatGPT, the landscape of disinformation generation and detection has changed. Disinformation generated by machines is evolving through the rapid development of advanced large language models. In this study, we focus on the novel research problem of detecting evolving LLM-generated disinformation (right-bottom square).}
% \vspace{-.3cm}
\label{fig:overview}
\end{figure}

Technically, the detection of disinformation generated by LLMs presents unique challenges. Traditional methods focused on human-written disinformation, including disinformation, conspiracy theories, and propaganda~\cite{shu2017fake}. These methods relied on linguistic cues and inconsistencies that are often present in human-generated content~\cite{zhou2020survey, zhang2020overview}. However, LLM-generated disinformation is inherently more complicated and harder to detect. These models can produce content that is not only coherent and contextually relevant but also tailored to specific audiences, making it more persuasive and difficult to identify as false~\cite{zhao2023more}.

There have been some initial works exploring LLM-generated disinformation detection~\cite{jiang2024disinformation, xu2023earth, sun2024exploring}. However, as LLMs are continually improving, their ability to mimic human writing styles and generate contextually persuasive disinformation is advancing rapidly too. Disinformation generated by LLMs is evolving alongside developing new LLM architectures, advanced training and alignment paradigms, and expanding model knowledge bases~\cite{chen2023combating}. To the best of our knowledge, no existing work addresses the dynamic and evolving nature of LLM-generated disinformation. Thus, as shown in Figure~\ref{fig:overview}, we aim to bridge this gap by formulating a new research problem: \textit{detecting evolving LLM-generated disinformation}.  

In this paper, we first demonstrate that current detection methods struggle with knowledge retention, as learning disinformation patterns from a long sequence of diverse LLMs often leads to significant knowledge loss and a decline in detection performance. Second, to overcome these challenges, we propose \textit{DELD}, a novel approach designed to enhance the detection of evolving LLM-generated disinformation. Our method leverages the general fact-checking capabilities of pre-trained language models and integrates the unique disinformation characteristics of various LLMs. By sequentially concatenating learned characteristics, \textit{DELD} facilitates the accumulation and transformation of model-specific knowledge, addressing the issue of knowledge loss. In particular, \textit{DELD} tackles the problem of label scarcity by combining semantic embeddings of news reports with trainable soft prompts, effectively eliciting model-specific knowledge.

Our extensive analysis and experimental results demonstrate that \textit{DELD} significantly outperforms state-of-the-art methods. Beyond improved detection accuracy, our method also provides critical insights into the unique patterns of disinformation generation across different LLMs, contributing valuable perspectives to this field of research. 

Our main contributions are summarized as follows:
\begin{itemize}[leftmargin=*]
    \item \underline{\textbf{Problem Formulation}}: We identify an urgent issue in the field of disinformation detection and formulate a novel research problem focused on detecting evolving disinformation generated by large language models.
    \item \underline{\textbf{Method Proposed}}: We propose \textit{DELD}, a simple and efficient framework that exploits both the fact-checking capabilities of language models and the model-specific knowledge of various disinformation generators to detect evolving disinformation.
    \item \underline{\textbf{Effectiveness Evaluation}}: We evaluate the effectiveness of \textit{DELD} through extensive experiments and analysis on real-world datasets. 
\end{itemize}

\begin{figure*}[!t]
\centering
\includegraphics[width=1\textwidth]{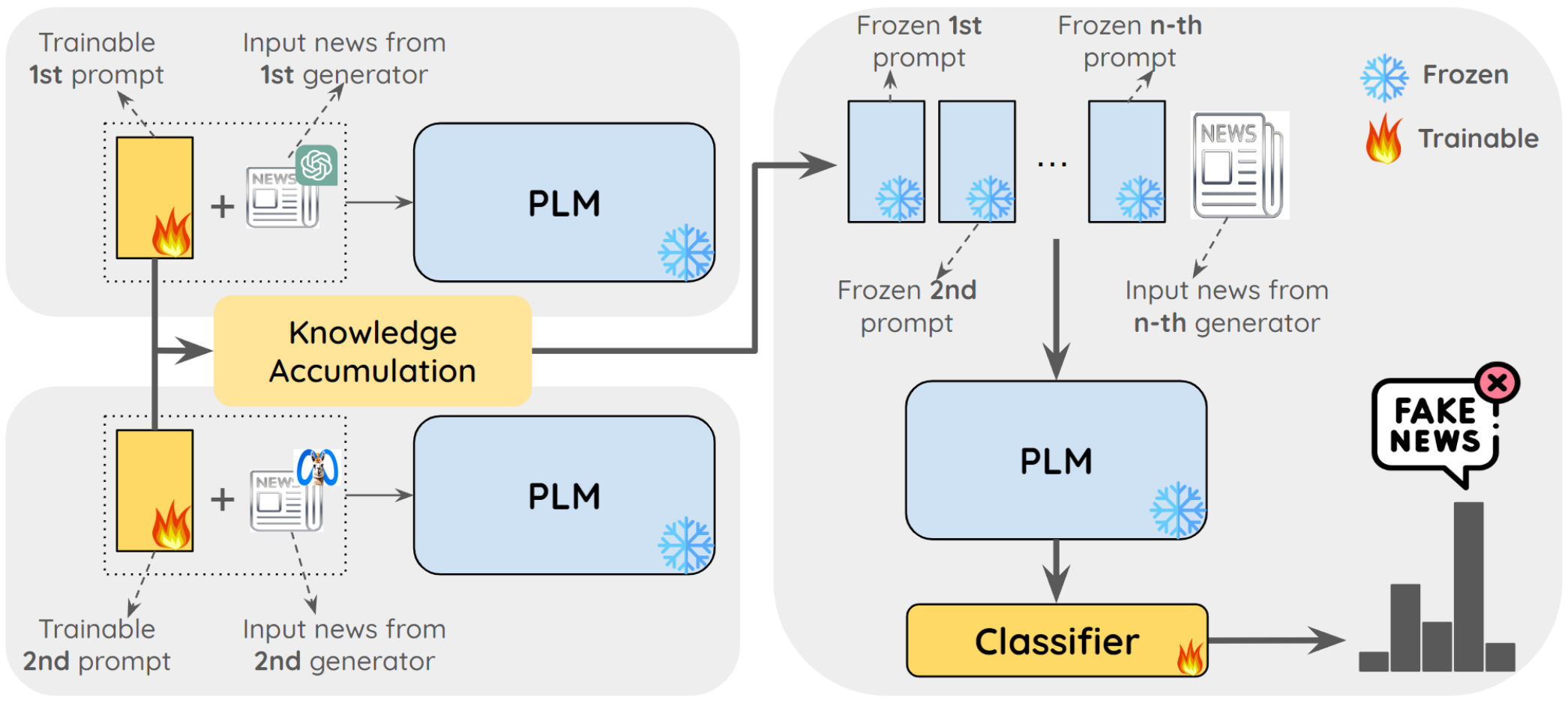} 
\caption{An overview of the proposed method \textit{DELD}, The left panel of the figure shows the training pipeline for trainable prompts. We train an individual soft prompt for each disinformation generator. Next, we concatenate them together to facilitate knowledge accumulation and transformation. On the right panel, which show after concatenating each prompt, we only finetune the classifier while keeping the learned prompts and the PLM frozen.}
\label{fig:method}
\end{figure*}
\section{Related Work}

\subsection{Disinformation Detection}
\textit{Disinformation}, refers to deliberately inaccurate or false information intended to deceive people, poses significant challenges in the digital age~\cite{shu2020disinformation}. The proliferation of disinformation on digital platforms makes manual fact-checking inefficient. Therefore, researchers have developed advanced automated techniques for disinformation detection~\cite{shu2017fake}.

\paragraph{Human-Written Disinformation Detection}
In recent years, the advent of deep learning has been a significant milestone in the field of disinformation detection~\cite{nasir2021fake}. Deep neural networks have been trained on large text corpora using a wide range of architectures such as Convolutional Neural Networks (CNNs)~\cite{zhang2015character} and Recurrent Neural Networks (RNNs)~\cite{przybyla2020capturing} to learn various textual features including semantic patterns and stylistic nuances~\cite{wu2023decor}. For example, Ruchansky et al.~\cite{ruchansky2017csi} built a hybrid model to capture news content, social networks, and temporal information to enhance disinformation detection. FakeBERT~\cite{kaliyar2021fakebert} utilizes Bidirectional Encoder Representations from Transformers (BERT) with several blocks of a single-layer CNN, combining with various kernel sizes and filters to identify disinformation. Jeong et al.~\cite{jeong2022nothing} proposed a Graph Neural Network (GNN) based model to extract the high-order relations of disinformation propagation and user interaction on social media.

\paragraph{LLM-Generated Disinformation Detection}
With the unprecedented text understanding and generation capability of LLMs, detecting disinformation generated by LLMs has attracted increasing attention~\cite{spitale2023ai}. Initial analyses have revealed that LLM-generated disinformation tends to be more convincing than those written by humans~\cite{zhou2023synthetic}. LLMs not only preserve the core falsehoods but also improve the narratives with more persuasive and formal language. Chen et al.~\cite{chen2023can} pointed out that it is harder for human experts to detect LLM-generated disinformation than human-generated ones. Jiang et al.~\cite{jiang2024disinformation}  demonstrated that models trained solely on datasets of human-written disinformation are less effective at detecting LLM-generated disinformation. Moreover, recent works explored the detection of AI-generated disinformation across multiple domains such as Politics, Health, and Science~\cite{nanabala2024unmasking, xu2024future}.

In the real world, the fast iteration of LLMs and their variants makes LLM-generated disinformation detection become more challenging. In this context, we investigate the novel problem of evolving LLM-generated disinformation detection. We focus on providing effective methods that are applicable to the constantly evolving nature of LLM-generated disinformation, ensuring robustness against future disinformation.

\subsection{Parameter-Efficient Fine-Tuning}
Parameter-efficient fine-tuning (PEFT) has emerged as a significant approach in the adaptation of large pre-trained models to specific tasks while minimizing the computational and storage overhead. Traditional fine-tuning methods involve updating all model parameters, which can be computationally expensive and impractical, especially for LLMs with billions of parameters~\cite{min2023recent,bender2021dangers}. Moreover, they often lead to the catastrophic forgetting problem~\cite{kirkpatrick2017overcoming}. PEFT addresses these challenges by fine-tuning only a small subset of the model's parameters, thereby achieving task-specific adaptation with reduced resource requirements. Several techniques have been proposed to achieve parameter efficiency in fine-tuning~\cite{houlsby2019parameter}. For example, \textit{prompting} methods such as \textit{soft prompt}~\cite{lester2021power}, \textit{prefix tuning}~\cite{li2021prefix}, and \textit{P-tuning}~\cite{liu2023gpt} involve learning a set of continuous prompt embeddings, which are prepended to the input sequence. Unlike traditional discrete prompts, soft prompts are trainable vectors optimized during the training process, while the original model parameters remain fixed. Moreover, \textit{Adapters} such as \textit{LoRA}~\cite{hu2021lora}, \textit{QLoRA}~\cite{dettmers2024qlora}, and \textit{LLaMA-Adapter}~\cite{zhang2023llama}, are small bottleneck layers inserted within each layer of the pre-trained language model. During fine-tuning, only the parameters of these adapter layers are updated while the rest of the model remains unchanged. Both of them offer efficient alternatives to traditional fine-tuning, enabling effective and resource-efficient adaptation of large language models to a wide range of tasks.

In the context of detecting evolving LLM-generated disinformation, PEFT methods offer several benefits. They allow for the continuous adaptation of detection models to new types of disinformation generated by evolving LLMs without the need for extensive computational resources. This is particularly crucial given the rapid pace at which LLMs are developed and deployed.

% \subsection{Continual Learning in NLP}
% \textit{Continual learning (CL)} is designed for models to achieve human-like knowledge learning, accumulating, and transferring capability~\cite{parisi2019continual}. However, when training with a continuous data stream, the model needs to adjust the network parameters, which may cause catastrophic forgetting (CF) of previously acquired knowledge~\cite{mccloskey1989catastrophic}. Existing techniques for controlling CF can be generally categorized into three groups: \textit{rehearsal-based}, \textit{regulation-based}, and \textit{architecture-based} approaches~\cite{biesialska2020continual}. Rehearsal-based methods save a subset of training samples of each previous task in a memory buffer and periodically replay them while training the model~\cite{rebuffi2017icarl}. Regularization-based CL is achieved by introducing a penalty or regularization to the loss function to penalize changes to important parameters learned from previous tasks when learning a new task~\cite{kirkpatrick2017overcoming}. Architecture-based techniques allocate a subset of task-specific parameters to each task to prevent CF~\cite{rusu2016progressive}. 

\section{Problem Formulation}
Similar to previous work~\cite{shu2017fake, shu2019defend}, we treat the detection of evolving LLM-generated disinformation as a binary classification problem. Denote $D = \{D_1, D_2, \dots, D_{k}\}$ as a set of LLM-generated disinformation datasets. Let the training set for the \( k \)-th disinformation generator \( D_k \) be \(\{(X_{i}, Y_{i})\}_{i=1}^{N}\), where $X_{i}$ represents a news article, $Y_{i} \in \{0, 1\}$ is the class label (0 = true news, 1 = disinformation), and $N$ is the sample size of $D_k$. Overall, we want to predict the class label $Y_{i}$ for a given news article $X_{i}$. Mathematically, we aim to learn the mapping function: $X \to Y$.

We assume that the model is parameterized by $\Theta$ and has access to the disinformation generator identity during training but not during inference. Hence, the learning objective across all disinformation generators becomes:

\begin{equation}
    \max_{\Theta} \hspace{0.15cm} \sum\limits_{n=1}^{k} \hspace{-0.05cm} \sum\limits_{X, Y \in D_k} \hspace{-0.1cm} \log p_{\Theta}(Y \mid X)
\end{equation}

Given these notations, we sequentially optimize the loss for the \( k \)-th generator by updating all model parameters:

\begin{equation}
\begin{aligned}
    \mathcal{L}_k(\Theta) = - \hspace{-0.25cm} \sum\limits_{X,Y \in D_k} \hspace{-0.2cm} \log p(Y \mid X, \Theta)
\end{aligned}
\end{equation}

\section{Proposed Method}
In this section, we introduce the proposed method \textit{DELD}, which is designed to effectively detect evolving disinformation generated by LLMs. Our method leverages parameter-efficient fine-tuning by integrating semantic embeddings of news articles with trainable soft prompts specific to each disinformation generator. We utilize BERT~\cite{devlin2018bert} and T5~\cite{raffel2020exploring} as the pre-trained language model, freezing its parameters while tuning only the soft prompts. This approach ensures computational efficiency and the ability to adapt to salient disinformation characteristics for different generators.

\subsection{Architecture Overview}
As illustrated in Figure~\ref{fig:method}, the architecture of \textit{DELD} consists of three main components:
\begin{itemize}
    \item \textbf{Pre-trained Language Model}: We employ a pre-trained language model as the backbone of our framework. The parameters of PLM are kept frozen during training to maintain the pre-trained knowledge and ensure training efficiency.
    \item \textbf{Soft Prompts}: Soft prompts are introduced as additional parameters that are trainable for each disinformation generator. The trained prompts are concatenated sequentially to facilitate knowledge accumulation.
    \item \textbf{Feature Integration}: Semantic embeddings of news articles are extracted and appended to the soft prompts to enhance the model's understanding of disinformation. With the aggregated representation, we train a classifier to detect disinformation.
\end{itemize}

% \subsection{Proposed Framework - \textit{DELD}}
\textit{DELD} adopts a straightforward prompt learning strategy. For each disinformation dataset (i.e., disinformation generated by a specific LLM), a unique soft prompt is learned and concatenated with previously learned soft prompts. This approach helps in preserving the knowledge learned from previous datasets and promotes forward transfer to unseen datasets.

Let $\mathbf{X} \in \mathbb{R}^{n \times d}$ represent the input sequence of news tokens, where $n$ is the sequence length and $d$ is the embedding dimension. The pre-trained language model, denoted as $\text{PLM}(\cdot)$, processes this input sequence to extract the semantic embedding. For each disinformation dataset $D_k$ (where $k \in \{1, \ldots, m\}$), we introduce a trainable prompt vector $\mathbf{P}_k \in \mathbb{R}^{m \times d}$, where $m$ is the number of prompt tokens for dataset $k$.

The input to the frozen PLM for dataset $D_k$ is the concatenation of all learned prompts $\mathbf{P}_i$ (for $i \leq k$) and the extracted semantic embedding:
\begin{equation}
    \mathbf{X'} = [\mathbf{P}_k; \mathbf{P}_{k-1}; \ldots; \mathbf{P}_1; \mathbf{X}] ,
\end{equation}
where $\mathbf{X'} \in \mathbb{R}^{(km+n) \times d}$. The output of the frozen PLM for the input sequence $\mathbf{X'}$ is given by:
\begin{equation}
    \mathbf{H} = \text{PLM}(\mathbf{X'}),
\end{equation}
where $\mathbf{H} \in \mathbb{R}^{(km+n) \times d}$ represents the hidden states of the aggregated representation.
\\
\\
\noindent\textbf{Training Objective:} For each dataset $D_k$, the objective is to find the prompt parameters $\theta_{\mathbf{P}_k}$ that minimize the negative log probability of the training examples under the soft prompts and the frozen pre-trained model:
\begin{equation}
% \small
\mathcal{L}(\theta_{\mathbf{P}_k}) = - \hspace{-0.35cm} \sum_{X, Y \in D_k} \hspace{-0.3cm} \log p(y \mid [\mathbf{P}_k, \ldots, \mathbf{P}_1, x], \theta, \theta_{\mathbf{P}_1}, \ldots, \theta_{\mathbf{P}_k}),
\end{equation}
% \begin{equation}
% \begin{aligned}
% \mathcal{L}(\theta_{\mathbf{P}_k}) = - \sum_{X, Y \in D_k} & \log p(y \mid [\mathbf{P}_k, \ldots, \mathbf{P}_1, x], \\
% & \theta, \theta_{\mathbf{P}_1}, \ldots, \theta_{\mathbf{P}_k}),
% \end{aligned}
% \end{equation}
% \begin{equation}
% \mathcal{L}(\theta_{\mathbf{P}_k}) = - \sum \log p(y \mid [\mathbf{P}_k, \ldots, \mathbf{P}_1, x], \theta, \theta_{\mathbf{P}_1}, \ldots, \theta_{\mathbf{P}_k}),
% \end{equation}
where $\theta$ are the frozen parameters of the PLM model and $\theta_{\mathbf{P}_i}$ are the parameters of the soft prompts learned for previous disinformation datasets. This setup allows \textit{DELD} to achieve two primary goals:
\begin{itemize}
    \item \textit{Eliminating Catastrophic Forgetting}: Since each soft prompt is trained independently and is frozen after the current dataset is learned, previously learned data do not suffer from forgetting.
    \item \textit{Knowledge Sharing}: Soft prompts learned on previous datasets facilitate information reuse, benefiting the learning of new datasets.
\end{itemize}

\noindent\textbf{Classification:} The final hidden states $\mathbf{H}$ are used for classification. Specifically, we apply a linear transformation followed by a sigmoid activation function to predict the likelihood of the input sequence being disinformation:
\begin{equation}
\hat{y} = \sigma(\mathbf{W} \mathbf{H} + \mathbf{b})
\end{equation}
where $\mathbf{W} \in \mathbb{R}^{k \times d}$ and $\mathbf{b} \in \mathbb{R}^k$ are the weights and biases of the classification layer, $k$ is the number of classes, and $\sigma(\cdot)$ denotes the sigmoid activation function for binary classification.

By focusing on the fine-tuning of soft prompts and integrating semantic embeddings, our proposed method \textit{DELD} effectively adapts to the evolving nature of LLM-generated disinformation. This prompt learning-based strategy ensures superior performance with minimal computational overhead, addressing the challenges of catastrophic forgetting and promoting forward knowledge transferring.

\subsection{Complexity Analysis}
We analyze the computational complexity of our proposed method, \textit{DELD}, focusing on both time complexity and space complexity.

\noindent\textbf{Time Complexity:} The overall time complexity of \textit{DELD} can be decomposed into the following components:
\begin{itemize}
    \item \textit{Pre-trained Model Inference}: The time complexity of a forward pass through PLM is $O(n^2 \cdot d \cdot L)$, where $n$ is the input sequence length, $d$ is the hidden dimension, and $L$ is the number of transformer layers. Since the parameters of PLM are frozen and not updated during training, this complexity remains constant for each input sequence.
    \item \textit{Soft Prompts Concatenation}: For each task $T_k$, we concatenate the soft prompts $\mathbf{P}_1, \mathbf{P}_2, \ldots, \mathbf{P}_k$ with the input sequence. This step involves copying the embeddings and has a linear complexity $O(k \cdot m \cdot d)$, where $k$ is the current task index and $m$ is the number of prompt tokens.
\end{itemize}

Combining these components, the overall time complexity for processing a single input sequence during training is:
\begin{equation}
O(n^2 \cdot d \cdot L + k \cdot m \cdot d)
\end{equation}
Given that $k \leq m$, this can be simplified to:
\begin{equation}
O(n^2 \cdot d \cdot L + m^2 \cdot d)
\end{equation}
where the quadratic term $n^2 \cdot d \cdot L$ dominates for long input sequences.
\\
\\
\noindent\textbf{Space Complexity:} The space complexity of \textit{DELD} primarily involves the storage of model parameters and intermediate representations:
\begin{itemize}
    \item \textit{Pre-trained Model Parameters}: The frozen PLM model has a fixed number of parameters, denoted as $|\theta|$, which does not change during training. This contributes a constant space complexity of $O(|\theta|)$.
    \item \textit{Soft Prompts Storage}: For each dataset $D_k$, we store a separate soft prompt $\mathbf{P}_k$ with $m$ prompt tokens, each of dimension $d$. The total space required for all soft prompts across $m$ datasets is:
    \begin{equation}
    O(m \cdot m \cdot d) = O(m^2 \cdot d)
    \end{equation}
    \item \textit{Intermediate Representations}: During inference, the hidden states $\mathbf{H}$ are stored. The space complexity for these representations is:
    \begin{equation}
    O((n + m \cdot m) \cdot d) = O((n + m^2) \cdot d)
    \end{equation}
\end{itemize}
Combining these components, the overall space complexity is:
\begin{equation}
O(|\theta| + m^2 \cdot d + (n + m^2) \cdot d)
\end{equation}
This can be simplified to:
\begin{equation}
O(|\theta| + m^2 \cdot d + n \cdot d)
\end{equation}

In summary, the computational complexity of our proposed method \textit{DELD} is primarily dominated by the inference complexity of the pre-trained model and the storage requirements for the soft prompts and intermediate representations. The time complexity is $O(n^2 \cdot d \cdot L + m^2 \cdot d)$, and the space complexity is $O(|\theta| + m^2 \cdot d + n \cdot d)$. These complexities highlight the efficiency of \textit{DELD} in adapting to evolving disinformation while maintaining computational feasibility.

\section{Experiments}
In this section, we present the experiments to evaluate the effectiveness of the proposed \textit{DELD} framework. Specifically, we aim to answer the following research questions:
\begin{itemize}
    \item Can conventional methods detect evolving LLM-generated disinformation?
    \item If not, can LLMs themselves be adapted to detect such disinformation?
    \item How effective is \textit{DELD} in detecting evolving LLM-generated disinformation?
    % \item Is \textit{DELD} robustness enough against different scenarios?
\end{itemize}

\subsection{Datasets}
We utilize an recent LLM-generated disinformation dataset~\cite{chen2023can}, which includes a variety of disinformation generated by state-of-the-art LLMs such as ChatGPT~\cite{OpenAI2023GPT4TR}, LLaMA~\cite{touvron2023llama}, and Vicuna~\cite{chiang2023vicuna}. This dataset provides a diverse and comprehensive range of disinformation examples, enabling extensive experimentation to address our research questions. In addition, we incorporate a human-written disinformation dataset~\cite{verma2021welfake} to better simulate the evolving nature of real-world disinformation. By combining these datasets, we create a practical and representative experimental dataset. Detailed statistics of this final dataset are presented in Table~\ref{tab:dataset}.

% We utilize a comprehensive LLM-generated disinformation dataset~\cite{chen2023can} which encompasses a variety of disinformation generated by ChatGPT~\cite{OpenAI2023GPT4TR}, LLaMA~\cite{touvron2023llama}, and Vicuna1.3-7B~\cite{chiang2023vicuna}. This dataset provides sufficient sources for us to conduct various experiments to answer the research questions. Besides, we also include a human-written disinformation dataset~\cite{verma2021welfake} to simulate the evolving nature of disinformation better. We divide and conquer all the data to construct a final dataset for our experiment. The detailed statistics of the dataset are shown in Table~\ref{tab:dataset}.

\begin{table}
    \centering
    \caption{Datasets statistics.}
    \scalebox{1}{
    \begin{tabular}{lccccc}
         \hline\toprule
         \textbf{Dataset}&\text{$D_\textbf{human}$} & \text{$D_\textbf{vicuna}$} & \text{$D_\textbf{llama}$} & {$D_\textbf{chatgpt}$} & {Total} \\
         % \hline
         \midrule
         % Time period & 01/2020-01/2022 & 01/2020-01/2022\\
         % \hline
         \# true news & 2,500 & 500 & 501 & 501 & 4002\\ 
         % \# of true news & 21,417 & 21,417 & 1,000 & 1,000 \\
         % avg \# of words & 557& 208& 196& 151\\ \hline 
         % \midrule/
         \# disinformation & 2,500 & 326 & 557 & 587 & 3970\\ 
         % % \# of true news & 21,417 & 21,417 & 1,000 & 1,000 \\
         % headline & \checkmark & \checkmark &  &  \\ \hline
         % content & \checkmark & \checkmark & \checkmark & \checkmark \\

         % Politics & 6,846 & 6,760 \\
         % Left News & 4,462 & 4,452 \\
         % Gov. News & 1,574 & 1,568 \\
         % U.S. News & 797 & 747 \\
         % Middle-East News & 793 & 746 \\
         % Total & 23,525 & 23,278 \\
         \bottomrule \hline
    \end{tabular}}
    % \vspace{-0.4cm}
    \label{tab:dataset}
\end{table}

\subsection{Experimental Settings}
We consider the following disinformation detection methods as baselines for comparison with \textit{DELD}:

\begin{itemize}
    \item \textbf{LLaMA:} We use the pre-trained LLaMA2-7B to detect disinformation zero-shot.
    \item \textbf{ChatGPT:} ChatGPT (GPT-4-turbo) from OpenAI is utilized to detect disinformation zero-shot.
    \item \textbf{FT-All}: Fine-tuning a model on all data. 
    \item \textbf{FT-Per}: Fine-tuning a separate model per dataset.
    \item \textbf{FT-Seq:} Fine-tuning a model on a sequence of datasets. 
    % \item \textbf{DELD-per:} train a separate \textit{DELD} per dataset. 
    
    % \item \textbf{IndFT:} train a specific model for each dataset. This method is served as the upper bound. 
\end{itemize}

For the zero-shot detection method LLaMA and ChatGPT, we use a prompt template as shown in Figure~\ref{fig:prompt}. We consider FT-All and FT-Per to be the ideal cases (i.e., upper bound) for detecting disinformation. FT-Seq is designed to simulate the real-world scenario in which evolving disinformation is appearing on the internet in sequential order. We select two representative PLMs BERT-base and T5-base, to illustrate our method is model-agnostic.
\begin{figure}[!tbh]
    \centering
    \begin{tcolorbox}[colback=yellow!10!white,colframe=gray!10!black,title=\text{\text{Prompt for Zero-Shot Disinformation Detection}:}]
    \textbf{System Prompt:}\\
    Act as a disinformation detector. Given the following news piece, which category does this news belong to? Return ``1'' if you think the news piece is disinformation; otherwise, return ``0''. Note that there is no need for an explanation.
    % 5. According to step 4, 
    % 5. Please respond with your confidence score on a scale of 1 to 100.\\}
    \tcblower
    \textbf{Context Prompt:}\\
    news: [\textit{the given news piece $\mathbf{X}$}]
    \end{tcolorbox}
    \caption{The prompt template for LLaMA and ChatGPT.}
    \label{fig:prompt}
    % \vspace{-0.5cm}
\end{figure}

\subsection{Implementation Details}
All experiments were conducted on an NVIDIA A100-SXM4-80GB GPU. \textit{DELD} is a model-agnostic method that can be injected into any pre-trained language model. The training aimed to fine-tune the soft prompts while maintaining the frozen parameters of the pre-trained model. In this work, we use the BERT-base and T5-base models from HuggingFace Transformers as the backbone. The learning rate is set to 1e-3. We fine-tune the models for 10 epochs. We set the soft prompt length to 4, 8, 12, 16, and 20 tokens. We randomly choose 80\% news pieces for training and the remaining 20\% for testing; the process is performed 10 times, and the average performance is reported. 

\subsection{Main Results}
In this section, we present the performance of \textit{DELD} in detecting LLM-generated disinformation. The results are compared against several baselines, including models without training, models trained on all data, models trained on individual datasets, and models trained sequentially. The evaluation metrics are based on the accuracy (\%) of disinformation detection across different datasets: $D_\text{human}$, $D_\text{vicuna}$, $D_\text{llama}$, and $D_\text{chatgpt}$.

\begin{table}
    \centering
    \caption{Performance comparison of various disinformation detection models. \textit{LLaMA} and \textit{ChatGPT} denote models without any fine-tuning. \textit{FT-All} refers to fine-tuning on all datasets, \textit{FT-Per} denotes fine-tuning separate models per dataset, and \textit{FT-Seq} indicates fine-tuning on sequential datasets ($D_\text{human}$ $\rightarrow$ $D_\text{vicuna}$ $\rightarrow$ $D_\text{llama}$ $\rightarrow$ $D_\text{chatgpt}$). \textit{w/ DELD} represent baseline models with our method applied to detect various disinformation. Accuracy (\%) for all models and datasets is shown, with higher values indicating better performance. We also provide the average accuracy across all four datasets for each model.}

    \scalebox{1}{
    \begin{tabular}{lccccccc}
    \hline
         \toprule
           
         \multicolumn{1}{c}{} & \multicolumn{1}{c}{\text{$D_\text{human}$}} & \multicolumn{1}{c}{\text{$D_\text{vicuna}$}} & \multicolumn{1}{c}{\text{$D_\text{llama}$}} & \multicolumn{1}{c}{\text{$D_\text{chatgpt}$}} & \multicolumn{1}{c}{\text{Average}}\\
         \midrule
% zero shot
         \multicolumn{1}{l}{} & \multicolumn{5}{l}{\textbf{Zero-Shot (No FT)}}\\
            
         \multicolumn{1}{l}{\text{LLaMA}} & \multicolumn{1}{c}{33.51} & \multicolumn{1}{c}{24.71} & \multicolumn{1}{c}{31.46} & \multicolumn{1}{c}{20.09} & \multicolumn{1}{c}{27.44} \\
 
         \multicolumn{1}{l}{\text{ChatGPT}} & \multicolumn{1}{c}{10.26} & \multicolumn{1}{c}{31.93} & \multicolumn{1}{c}{48.36} & \multicolumn{1}{c}{43.84} & \multicolumn{1}{c}{33.61} \\   
         \midrule
% FT all
         \multicolumn{1}{c}{} & \multicolumn{5}{l}{\textbf{FT on all data}}\\

         \multicolumn{1}{l}{\text{FT-All (BERT)}} & \multicolumn{1}{c}{77.50} & \multicolumn{1}{c}{71.69} & \multicolumn{1}{c}{69.48} & \multicolumn{1}{c}{87.21} & \multicolumn{1}{c}{76.47} \\     
         \rowcolor{gray!30} % Add this line

         \multicolumn{1}{l}{\text{\hspace{2mm} w/ DELD}} & \multicolumn{1}{c}{94.42} & \multicolumn{1}{c}{80.64} & \multicolumn{1}{c}{81.94} & \multicolumn{1}{c}{92.45} & \multicolumn{1}{c}{87.36} \\         
         \multicolumn{1}{l}{\text{FT-All (T5)}} & \multicolumn{1}{c}{79.42} & \multicolumn{1}{c}{73.45} & \multicolumn{1}{c}{74.36} & \multicolumn{1}{c}{86.35} & \multicolumn{1}{c}{78.39} \\     
         \rowcolor{gray!30} % Add this line

         \multicolumn{1}{l}{\text{\hspace{2mm} w/ DELD}} & \multicolumn{1}{c}{96.12} & \multicolumn{1}{c}{83.35} & \multicolumn{1}{c}{85.24} & \multicolumn{1}{c}{92.10} & \multicolumn{1}{c}{89.20} \\            
         \midrule
% FT per
         \multicolumn{1}{c}{} & \multicolumn{5}{l}{\textbf{FT a model per dataset}}\\

         \multicolumn{1}{l}{\text{FT-Per (BERT)}} & \multicolumn{1}{c}{74.09} & \multicolumn{1}{c}{77.29} & \multicolumn{1}{c}{68.97} & \multicolumn{1}{c}{87.21} & \multicolumn{1}{c}{76.89} \\       
         \rowcolor{gray!30} % Add this line

         \multicolumn{1}{l}{\text{\hspace{2mm} w/ DELD}} & \multicolumn{1}{c}{\text{93.61}} & \multicolumn{1}{c}{89.52} & \multicolumn{1}{c}{\text{82.63}} & \multicolumn{1}{c}{91.78} & \multicolumn{1}{c}{\text{89.39}} \\    
         \multicolumn{1}{l}{\text{FT-Per (T5)}} & \multicolumn{1}{c}{77.30} & \multicolumn{1}{c}{77.38} & \multicolumn{1}{c}{69.20} & \multicolumn{1}{c}{91.37} & \multicolumn{1}{c}{78.81} \\      
         \rowcolor{gray!30} % Add this line

         \multicolumn{1}{l}{\text{\hspace{2mm} w/ DELD}} & \multicolumn{1}{c}{\text{98.75}} & \multicolumn{1}{c}{84.34} & \multicolumn{1}{c}{\text{89.77}} & \multicolumn{1}{c}{92.58} & \multicolumn{1}{c}{\text{91.36}} \\          
         \midrule         

% FT seq         
         \multicolumn{1}{c}{} & \multicolumn{5}{l}{\textbf{FT on sequential datasets}}\\

         \multicolumn{1}{l}{\text{FT-Seq (BERT)}} & \multicolumn{1}{c}{69.62} & \multicolumn{1}{c}{66.27} & \multicolumn{1}{c}{74.65} & \multicolumn{1}{c}{83.11} & \multicolumn{1}{c}{73.40} \\
         \rowcolor{gray!30} % Add this line

         \multicolumn{1}{l}{\text{\hspace{2mm} w/ DELD}} & \multicolumn{1}{c}{\textbf{82.71}} & \multicolumn{1}{c}{\textbf{81.93}} & \multicolumn{1}{c}{\textbf{78.87}} & \multicolumn{1}{c}{\textbf{97.26}} & \multicolumn{1}{c}{\textbf{85.19}} \\        
         \multicolumn{1}{l}{\text{FT-Seq (T5)}} & \multicolumn{1}{c}{72.53} & \multicolumn{1}{c}{71.84} & \multicolumn{1}{c}{77.31} & \multicolumn{1}{c}{86.10} & \multicolumn{1}{c}{76.95} \\
         \rowcolor{gray!30} % Add this line
         \multicolumn{1}{l}{\text{\hspace{2mm} w/ DELD}} & \multicolumn{1}{c}{\textbf{91.20}} & \multicolumn{1}{c}{\textbf{86.71}} & \multicolumn{1}{c}{\textbf{86.10}} & \multicolumn{1}{c}{\textbf{97.81}} & \multicolumn{1}{c}{\textbf{90.44}} \\     

         \bottomrule
         \hline
    \end{tabular}
    }
    \label{tab:ex_result}
% \vspace{-0.3cm}
\end{table}

Our method, \textit{DELD}, demonstrated superior performance across various experimental settings. The main findings from our experiments are summarized as follows:

\begin{itemize}
    \item \textit{Zero-Shot (No FT)}: The models without any fine-tuning showed significantly lower performance across all datasets. Specifically, ChatGPT (GPT-4)  achieved an average accuracy of 33.61\%, while the LLaMA model achieved 27.44\%.
    \item \textit{Fine-tuning on All Data}: The FT-All setting, where the model was trained on the combined data from all datasets, resulted in a substantial improvement, with an average accuracy of 76.47\% for BERT and 78.39\% for T5. When enhanced with \textit{DELD}, the accuracy further increased, achieving 87.36\% and 89.20\% for BERT and T5, respectively.
    \item \textit{Fine-tuning a Model Per Dataset}: The FT-Per setting, which involves training a separate model for each dataset, achieved an average accuracy of 76.89\% for BERT and 78.81\% for T5. Our method, DELD-Per, significantly outperformed the FT-Per baselines, achieving the highest accuracy on the $D_\text{human}$ and $D_\text{llama}$ datasets, with an overall average accuracy of 89.39\% for BERT and 91.36\% for T5.
    \item \textit{Fine-tuning on Sequential Datasets}: The FT-Seq setting, which trains models sequentially on each dataset to simulate the evolving nature of LLM-generated disinformation, showed an average accuracy of 73.40\% for BERT and 76.95\% for T5. Our method achieved an average accuracy of 85.19\% for BERT and 90.44\% for T5, demonstrating its robustness and effectiveness in detecting evolving LLM-generated disinformation.
\end{itemize}

Overall, we observe that FT-All and FT-Per present relatively better detection accuracy, serving as upper bounds for comparison. These approaches assume access to all data simultaneously or the ability to fine-tune separate models for each dataset, which may not always be feasible in real-world scenarios. In contrast, FT-Seq, which sequentially fine-tunes models on each dataset, provides a more practical and scalable approach. 

Our results show that \textbf{traditional fine-tuning approaches and LLMs cannot detect evolving LLM-generated disinformation efficiently}. In contrast, \textit{DELD} consistently outperformed the baseline methods across different training configurations and datasets by a considerable margin. In particular, the results of FT-Seq with \textit{DELD} highlight \textbf{the effectiveness of our approach in adapting to the evolving nature of LLM-generated disinformation}.

\subsection{Impact of Prompt Design}
We explore the impact of prompt design on the performance of our proposed method for detecting LLM-generated disinformation. Specifically, we examine how the length of the prompts and their position (i.e., prepend vs. append) affect the model's accuracy. Note that all results for this subsection are from FT-Seq (BERT) with \textit{DELD}.

\textbf{Prompt Length:} We experimented with various prompt lengths, ranging from 4 to 20 tokens, and observed their impact on the model's performance. The results, shown in Table \ref{tab:prompt_position}, indicate that prompt length significantly affects the detection accuracy.

% \begin{table}[ht]
%     \centering
%     \caption{Impact of Prompt Length on Model Performance}
%     \scalebox{1}{
%     \begin{tabular}{lcccc}
%         \toprule
%         \textbf{Prompt Length} & \multicolumn{4}{c}{\textbf{Accuracy (\%)}} \\
%         \midrule
%         \textbf{1} & 91.9 & 85.2, 83.7 & 81.3, 76.5, 85.4 & 74.2, 78.3, 77.5, 94.1 & 81.0 \\
%         \textbf{2} & 93.2 & 84.9, 80.7 & 81.8, 84.9, 85.4 & 76.2, 78.3, 77.9, 96.3 & 82.2 \\
%         \textbf{3} & 93.6 & 88.8, 80.1 & 83.7, 82.5, 84.0 & 82.7, 81.9, 78.8, 97.3 & 85.2 \\
%         \textbf{4} & 94.5 & 83.2, 82.5 & 76.8, 79.5, 83.1 & 72.7, 78.9, 77.5, 95.4 & 81.1 \\
%         \textbf{5} & 94.1 & 85.4, 81.9 & 75.9, 80.7, 89.7 & 72.7, 77.7, 84.9, 89.5 & 81.2 \\
%         \bottomrule
%     \end{tabular}
%     }
%     \label{tab:prompt_length}
% \end{table}
From the results, we observe that:
\begin{itemize}
    \item \textit{4-Token Prompts}: Achieve moderate accuracy, with an average of 81.0\%.
    \item \textit{8-Token Prompts}: Show a slight improvement, with an average accuracy of 82.2\%.
    \item \textit{12-Token Prompts}: Perform the best overall, achieving the highest average accuracy of 85.2\%.
    \item \textit{16-Token and 20-Token Prompts}: Do not consistently improve performance, with average accuracies of 81.1\% and 81.2\%, respectively.
\end{itemize}

This indicates that a prompt length of 12 tokens strikes the best balance between providing sufficient context and maintaining simplicity.

\textbf{Prompt Position:} We also investigated the effect of the prompt position by comparing the default prepending method with appending. The performance results are summarized in Table \ref{tab:prompt_position}.

\begin{table}
    \centering
    \caption{Average accuracy (\%) of \textit{DELD} over different Prompt lengths and positions.}
    \scalebox{1}{
    \begin{tabular}{lcc}
    \hline
        \toprule
        \textbf{Prompt Length} & \textbf{Prepend} & \textbf{Append} \\
        \midrule
        \text{4} & 81.01 & 78.04 \\
        \text{8} & 82.19 & 78.65 \\
        \rowcolor{gray!30}
        \text{12} & \textbf{85.19} & \textbf{84.04} \\
        \text{16} & 81.13 & 79.66 \\
        \text{20} & 81.22 & 80.48 \\
        \bottomrule
        \hline
    \end{tabular}
    }
    \label{tab:prompt_position}
\end{table}

The results indicate that:
\begin{itemize}
    \item Prepending prompts generally outperforms appending prompts across all lengths.
    \item The most significant performance difference is observed with 12-token prompts, where prepending yields an accuracy of 85.2\% compared to 84.0\% for appending.
\end{itemize}

These findings suggest that placing prompts at the beginning of the input sequence (prepending) helps the model utilize the prompt information more effectively, possibly because it provides immediate context that influences the model's processing of the subsequent tokens. Overall, the analysis of prompt length and position reveals that \textbf{a 12-token prompt prepended to the input sequence delivers the best performance} for detecting LLM-generated disinformation. These insights can guide future research and applications in optimizing prompt design for various NLP tasks.

\subsection{Robustness to Training Order}
We examine the robustness of our proposed method, \textit{DELD}, against different orders of dataset presentation during training. Ensuring that the method is not biased or overly sensitive to the sequence in which data is presented is crucial for its applicability in real-world scenarios, where the order of data might vary. We tested our method on four different dataset orders and compared the performance to the default order used in the main results.

The following dataset orders were evaluated:
\begin{itemize}
    \item \textit{Order-1} (Default): $D_\text{human}$ $\rightarrow$ $D_\text{vicuna}$ $\rightarrow$ $D_\text{llama}$ $\rightarrow$ $D_\text{chatgpt}$
    \item \textit{Order-2}: $D_\text{vicuna}$ $\rightarrow$ $D_\text{human}$ $\rightarrow$ $D_\text{chatgpt}$ $\rightarrow$ $D_\text{llama}$
    \item \textit{Order-3}: $D_\text{llama}$ $\rightarrow$ $D_\text{chatgpt}$ $\rightarrow$ $D_\text{human}$ $\rightarrow$ $D_\text{vicuna}$
    \item \textit{Order-4}: $D_\text{chatgpt}$ $\rightarrow$ $D_\text{llama}$ $\rightarrow$ $D_\text{vicuna}$ $\rightarrow$ $D_\text{human}$
\end{itemize}

The results, summarized in Table \ref{tab:training_order}, indicate that while the order of datasets can influence performance to some extent, our method maintains robustness across different training orders. This consistency is essential for the following reasons:

\begin{itemize}
    \item \textit{Bias Mitigation}: The robustness against varying dataset orders suggests that the model does not overly rely on the specific sequence of data presentation. This characteristic helps in mitigating potential biases that could arise from a fixed data order.
    \item \textit{Generalization}: Maintaining high performance across different orders demonstrates the method's ability to generalize well. This implies that the model can effectively learn the underlying patterns of disinformation regardless of the order in which the data is presented.
    \item \textit{Practical Applicability}: In real-world applications, data is often collected and processed in a non-sequential manner. The ability of \textit{DELD} to remain effective under varying training conditions reinforces its practical utility and reliability.
\end{itemize}

From the results, it is evident that while there are minor variations in performance across different orders, these differences are not substantial enough to undermine the effectiveness of the proposed method. For instance, the performance with Order-1 (default) is slightly higher, but the method consistently achieves high accuracy across all other orders as well.

These findings reinforce the reliability and effectiveness of \textit{DELD}, \textbf{demonstrating its robustness against the variability in data presentation order}. This robustness is a critical attribute, ensuring that the method can be confidently applied in diverse and dynamic real-world environments where data sequences are not controlled or predictable.

\begin{table}
    \centering
    % \small
    \caption{Average accuracy (\%) of \textit{DELD} and FT-Seq over different training orders.}

    \scalebox{1}{
    \begin{tabular}{lcccccc}
    \hline
         \toprule
           
         \multicolumn{1}{c}{} & \multicolumn{1}{c}{\text{Order-1}} & \multicolumn{1}{c}{\text{Order-2}} & \multicolumn{1}{c}{\text{Order-3}} & \multicolumn{1}{c}{\text{Order-4}}\\
         \midrule
         % \multicolumn{6}{c}{\text{No training}}\\         

         % \multicolumn{6}{c}{\text{Train on sequential datasets}}\\

         \multicolumn{1}{l}{\text{FT-Seq}} & \multicolumn{1}{c}{69.62} & \multicolumn{1}{c}{67.06} & \multicolumn{1}{c}{78.12} & \multicolumn{1}{c}{73.63} \\
         \rowcolor{gray!30}
         \multicolumn{1}{l}{\text{\hspace{2mm} w/ DELD}} & \multicolumn{1}{c}{\textbf{85.19}} & \multicolumn{1}{c}{\textbf{83.78}} & \multicolumn{1}{c}{\textbf{87.11}} & \multicolumn{1}{c}{\textbf{83.37}} \\        

         \bottomrule
         \hline
    \end{tabular}
    }
    \label{tab:training_order}
\end{table}

\subsection{Evaluating Catastrophic Forgetting}
To evaluate catastrophic forgetting in our proposed method, we use the Forgetting (Fgt) metric. This metric quantifies the degree of forgetting of previous datasets after training on the last dataset. The degree of forgetting is calculated as follows:
\begin{equation}
    \text{Fgt} = \frac{1}{|D|-1} \sum_{i=1}^{|D|-1} \left( \max_{k=i}^{|D|-1} a_{i,k} - a_{i,|D|} \right)
    \label{eq:fgt}
\end{equation}
where $|D|$ is the total number of datasets, $a_{i,k}$ represents the accuracy (\%) on dataset $i$ after training on dataset $k$, and $a_{i,|D|}$ is the accuracy on dataset $i$ after training on the final dataset.

\begin{figure}[!t]
\centering
\includegraphics[width=1.1\columnwidth]{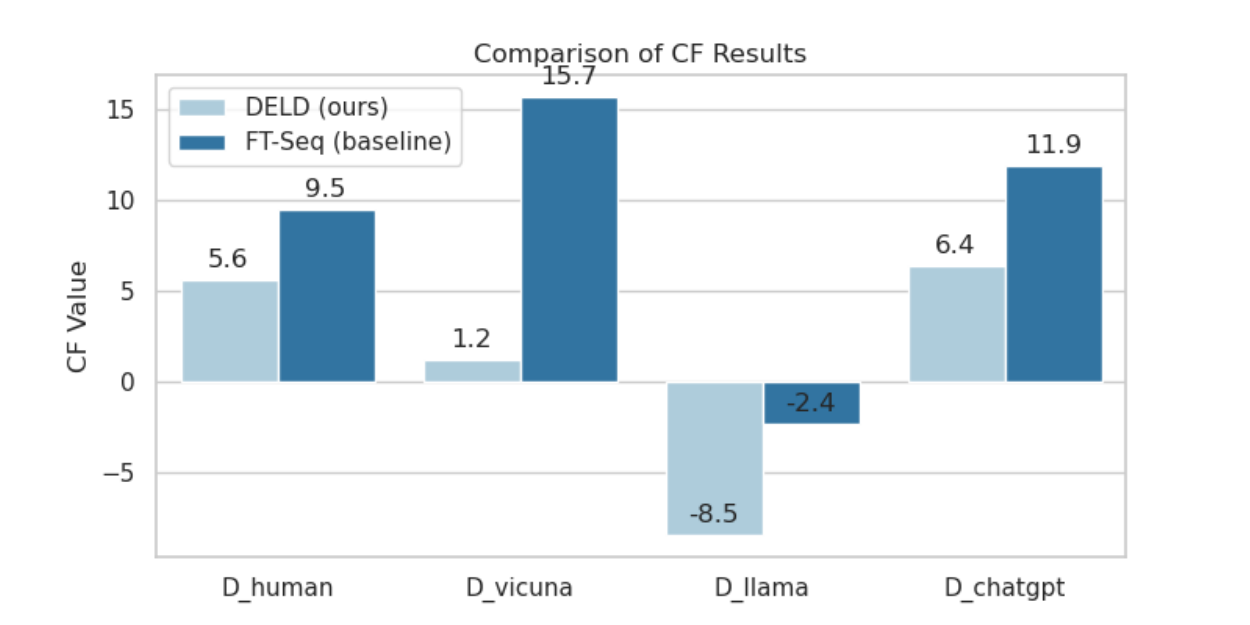} 
% \vspace{-.2cm}
\caption{Comparison of model forgetting across different datasets.}
% \vspace{-.3cm}
\label{fig:cf}
\end{figure}

We evaluated the effectiveness of these strategies in mitigating catastrophic forgetting using the Forgetting metric in Equation~\ref{eq:fgt}. The results, presented in Figure \ref{fig:cf}, show a substantial reduction in forgetting across various tasks compared to the baseline, indicating the robustness of our approach. Note that all results for this subsection are from FT-Seq (BERT) with \textit{DELD}.

The low values of the Forgetting (Fgt) metric across different datasets demonstrate the effectiveness of our method in mitigating catastrophic forgetting. Note that a negative forgetting value indicates a performance improvement after final training. We speculate this is because the model successfully transfers knowledge from other datasets. By sequentially concatenating the learned soft prompts, \textit{DELD} retains knowledge from previous datasets while adapting to new ones. These findings reinforce the practical viability of \textit{DELD} for detecting evolving LLM-generated disinformation that requires continual adaptation without sacrificing previously acquired knowledge.

\subsection{Characterize Disinformation Generator}
\begin{figure*}
    \centering \small
    \begin{subfigure}{0.45\textwidth}
        \includegraphics[width=1\columnwidth]{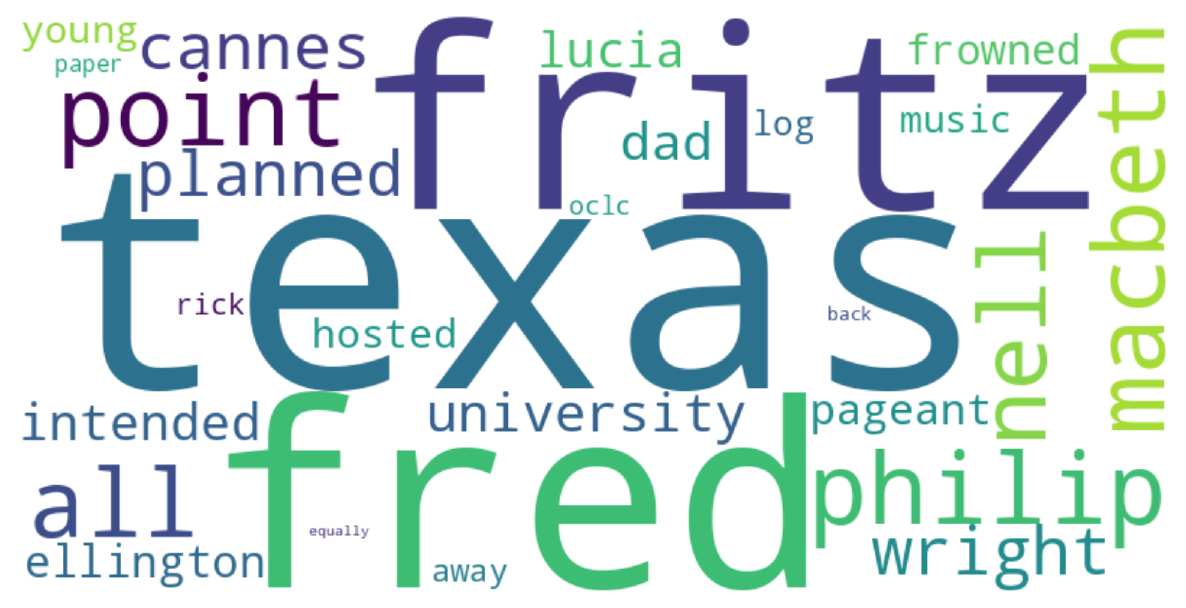}
        \caption{Human-written.}
        \label{fig:human}
    \end{subfigure}
    \begin{subfigure}{0.45\textwidth}
        \includegraphics[width=1\columnwidth]{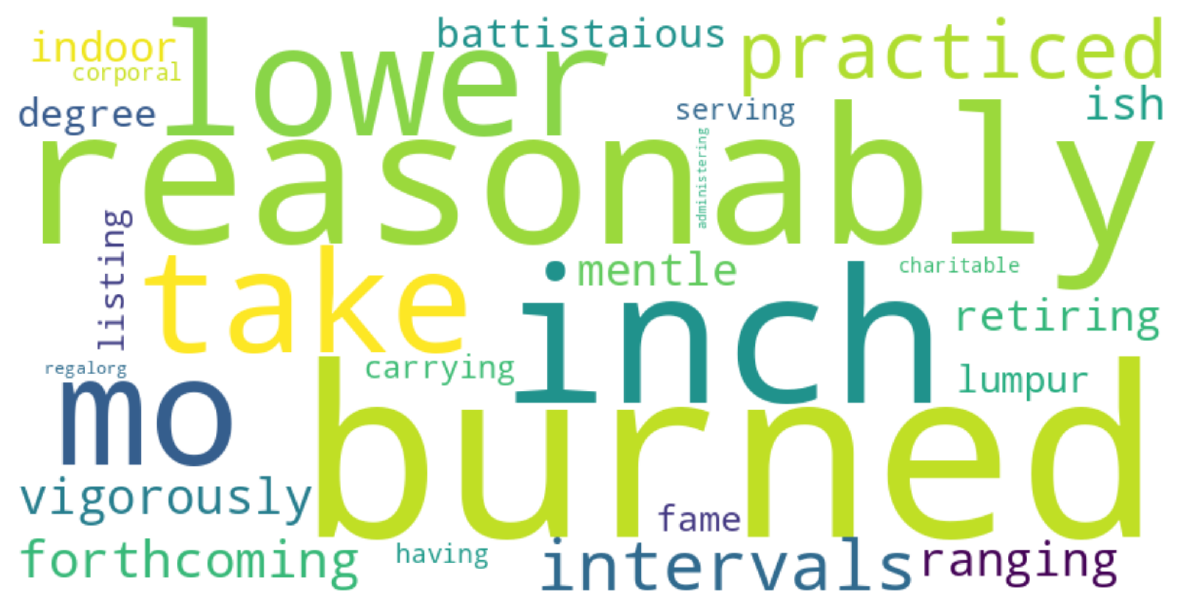}
        \caption{Vicuna.}
        \label{fig:vicuna}
    \end{subfigure}
    \begin{subfigure}{0.45\textwidth}
        \includegraphics[width=1\columnwidth]{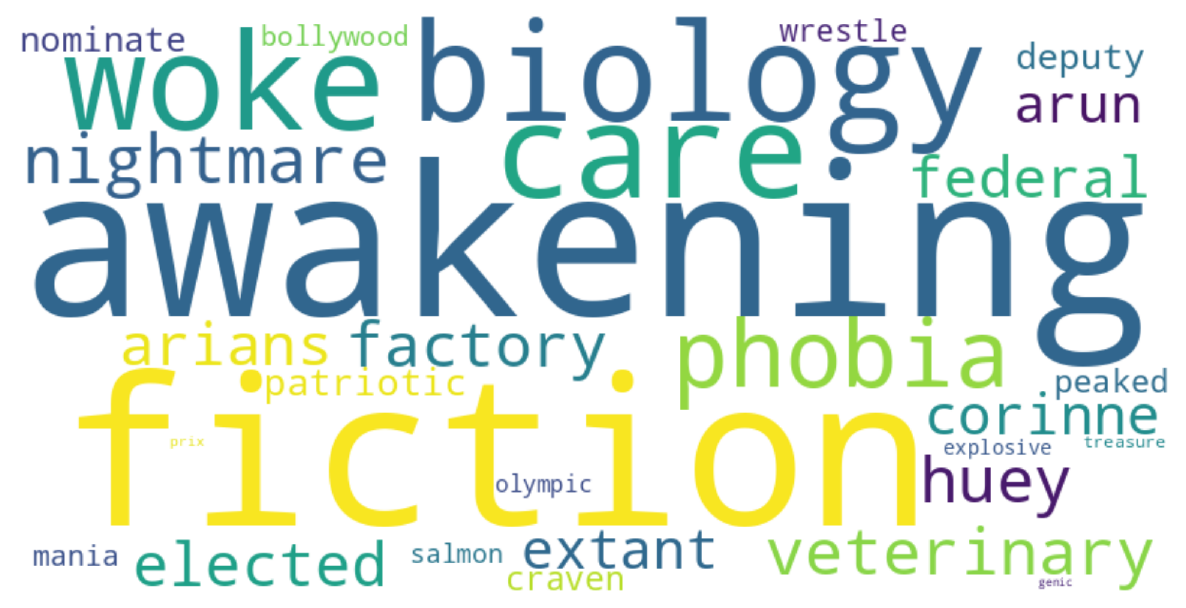}
        \caption{LLaMA.}
        \label{fig:llama}
    \end{subfigure}
    \begin{subfigure}{0.45\textwidth}
        \includegraphics[width=1\columnwidth]{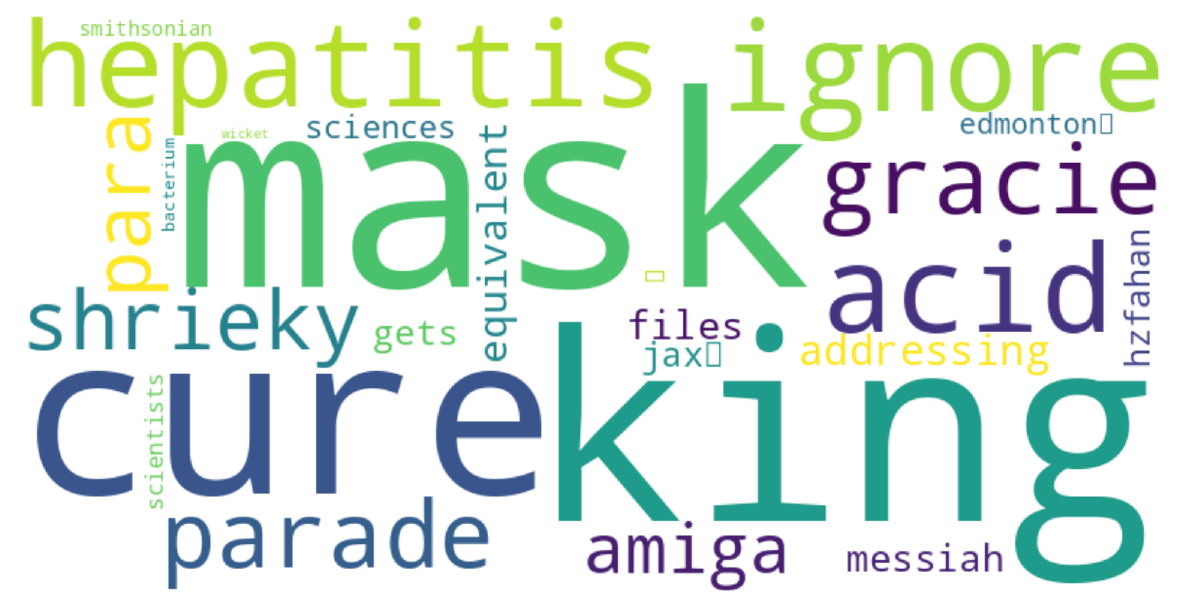}
        \caption{ChatGPT.}
        \label{fig:chatgpt}
    \end{subfigure}
    % \vspace{-.2cm}
        \caption{Illustration of characteristics of four disinformation generators using word cloud.}
    % \vspace{-0.2cm}
    \label{fig:wc}
\end{figure*}
We analyze the important tokens from the soft prompts used by various disinformation generators. Each set of words characterizes the distinct linguistic features associated with the disinformation generated by different models. 

\begin{itemize}
    \item \noindent\textit{Human-written:} As shown in Figure~\ref{fig:human}, the tokens from the human-generated disinformation dataset prompt reflect a broad and varied range of topics, including personal names, places, events, and general terms. This diversity highlights the unpredictable and wide-ranging nature of human-created disinformation.
    \item \noindent\textit{Vicuna:} In Figure~\ref{fig:vicuna}, disinformation generated by Vicuna is characterized by terms that suggest activities, events, and descriptive adjectives. This indicates a focus on dynamic scenarios and vivid descriptions, which can create engaging and credible narratives. It is likely designed to evoke specific images or actions, showing a tendency to construct stories that are action-oriented and rich in detail.
    \item \noindent\textit{LLaMA:} As illustrated in Figure~\ref{fig:llama}, the LLaMA model's disinformation covers a variety of domains, such as science, politics, and entertainment. This wide range of topics indicates the model's capability to generate content across multiple subjects, reflecting its relatively broad training data. Frequent technical and colloquial terms suggest that LLaMA can polish its disinformation to both specialized and general audiences, enhancing its versatility and potential impact.
    \item \noindent\textit{ChatGPT:} As presented in Figure~\ref{fig:chatgpt}, the tokens from the ChatGPT-specific prompt, which reveal a mix of medical terms, names, places, and general concepts. This points to the model's ability to generate disinformation on diverse topics, often blending factual-sounding details with false information. 
\end{itemize}

These important tokens from different disinformation generators highlight distinct patterns and characteristics for each model. Human-generated disinformation tends to be highly contextual and detailed. In contrast, the Vicuna, LLaMA, and ChatGPT models exhibit their own unique styles:
\begin{itemize}
    \item Vicuna focuses on dynamic and descriptive content, making its disinformation engaging and vivid.
    \item LLaMA covers more topics, indicating great versatility and adaptability to various contexts.
    \item ChatGPT combines medical, technical, and general terms to produce convincing and credible disinformation.
\end{itemize}

These differences highlight the importance of understanding the specific characteristics of each disinformation generator when developing detection and mitigation strategies.

\section{Discussion}
In this paper, we first point out a timely issue in the area of disinformation detection and formulate a novel research problem of detecting evolving LLM-generated disinformation. The proposed method, \textit{DELD}, demonstrates significant advancements in detecting evolving LLM-generated disinformation. Our approach leverages the general fact-checking capabilities of pre-trained language models and integrates the unique disinformation characteristics of various LLMs through the sequential concatenation of learned characteristics. This method addresses key challenges in the field, including disinformation knowledge accumulation and transformation, parameter-efficient fine-tuning, and catastrophic forgetting.

Our experiments reveal that \textit{DELD} significantly outperforms existing state-of-the-art methods. This improvement is attributed to the method's ability to preserve and transfer knowledge across different disinformation datasets, thereby enhancing the model's robustness and adaptability. The analysis of prompt design, particularly the impact of prompt length and position, further supports the effectiveness of our approach, with a 12-token prompt prepended to the input sequence achieving the best performance.

Moreover, the robustness of \textit{DELD} against different training orders indicates its practical applicability. In real-world scenarios, where data may be collected and processed in an unknown order, the ability of our method to maintain high performance across various training orders is crucial. Moreover, the method's efficiency in mitigating catastrophic forgetting highlights the potential of \textit{DELD} to be applied in diverse and dynamic environments.

The characterization of four disinformation datasets provides critical insights into the unique patterns of different disinformation generations. Human-generated disinformation tends to be highly contextual and detailed, whereas models like Vicuna, LLaMA, and ChatGPT exhibit distinct styles, focusing on dynamic and descriptive content, versatility across topics, and blending factual-sounding details with false information, respectively. Understanding these patterns is also essential for developing targeted detection and mitigation strategies for combating evolving LLM-generated disinformation.

\section{Conclusion and Future Work}
We formulate a novel research problem: detecting evolving LLM-generated disinformation. We proposed \textit{DELD}, a simple and efficient method that effectively leverages the fact-checking capabilities of pre-trained language models and integrates the unique disinformation characteristics of various LLMs. Our extensive experiments demonstrate that \textit{DELD} significantly outperforms existing methods, achieving superior performance in detecting evolving disinformation generated by different large language models.
We conduct substantial analysis to evaluate the robustness of \textit{DELD} against different training settings. Moreover, we present the efficiency of \textit{DELD} in mitigating catastrophic forgetting, highlighting its practical applicability in real-world scenarios. Additionally, the insights gained from analyzing the distinct characteristics of disinformation generators indicate the importance of understanding the specific patterns of each model when developing detection strategies. In the future, we plan to further enhance the robustness in more complex scenarios such as domain-shifts~\cite{tan2022graph}, noisy labels~\cite{wang2023noise}, attacks~\cite{Weng2024,weng2024big}, among others.

Overall, this paper contributes valuable perspectives to the field of disinformation detection and offers promising directions for future research. As LLMs continue to evolve, methods like \textit{DELD} will be crucial in maintaining the integrity of information and combating the growing threat of disinformation. In particular, many other disinformation settings are worth considering. For example, it is essential to develop an AI-generated dataset using various large language models and diverse prompts~\cite{mo2024large}. This dataset can be utilized to better simulate the real-world disinformation generation and detection process. Also, as LLMs start to show their great potential to generate high-quality images and videos, it is important to investigate evolving LLM-generated disinformation in a multi-modal scenario.

\section{Ethical Consideration}
The development and deployment of technologies for detecting LLM-generated disinformation raise several important ethical considerations that must be addressed to ensure responsible research and application~\cite{liu2024unraveling}. First, this work only focuses on publicly available disinformation datasets, which do not include any personal information. Second, while our goal is to combat disinformation, the technologies we develop could potentially be misused for harmful purposes, such as surveillance or censorship~\cite{jhaver2018online}. It is essential to consider the dual-use nature of our work and take steps to prevent misuse. We advocate for transparent reporting of our methods and findings, encourage the use of our technologies within ethical frameworks, and collaborate with policymakers to establish guidelines that prevent misuse. The implementation of disinformation detection systems can have significant effects on public discourse and freedom of expression. While it is important to reduce the spread of false information, it is equally important to ensure that legitimate content is not inadvertently censored~\cite{langvardt2017regulating}. It is important to balance these considerations by designing the detection systems to minimize false positives and by providing clear explanations for content flagged as disinformation. Engaging with stakeholders, including civil society organizations and the general public, helps us align our goals with societal values and expectations.

\bibliography{DSAA24}
\bibliographystyle{IEEEtran}
\end{document}